\documentclass[10pt,twocolumn,letterpaper]{article}
\newif\ifreview 
\newif\ifarxiv \newcommand{\arxiv}{\arxivtrue}
\newif\ifcamera 
\newif\ifrebuttal

\arxiv %
\ifreview \usepackage[review]{cvpr} \fi
\ifarxiv \usepackage[pagenumbers]{cvpr} \fi
\ifrebuttal \usepackage[rebuttal]{cvpr} \fi
\ifcamera \usepackage{cvpr} \fi

\usepackage[hang,flushmargin]{footmisc}
\usepackage{amsmath}
\usepackage{amssymb}
\usepackage{booktabs}
\usepackage{caption}
\usepackage{color}
\usepackage{colortbl}
\usepackage{enumitem}
\usepackage{epsfig}
\usepackage{float}
\usepackage{graphicx}
\usepackage{microtype}
\usepackage{multirow}
\usepackage{placeins}
\usepackage{stfloats}
\usepackage{subcaption}
\usepackage{tabularx}
\usepackage{times}
\usepackage{url}
\usepackage{xcolor}
\usepackage{xr-hyper}
\usepackage{xspace}
\usepackage{xstring}
\usepackage{textcomp}
\definecolor{cvprblue}{rgb}{0.21,0.49,0.74}
\definecolor{pinklink}{rgb}{0.98,0.44,0.57}
\usepackage[breaklinks,colorlinks,urlcolor=pinklink,linkcolor=cvprblue,citecolor=cvprblue]{hyperref}
\usepackage[capitalize]{cleveref}

\crefname{section}{Sec.}{Secs.}
\crefname{table}{Table}{Tables}
\crefname{figure}{Fig.}{Figs.}

\ifarxiv \crefname{appendix}{App.}{Apps.}
\else \crefname{appendix}{Suppl.}{Suppls.} \fi

\ifarxiv  \fi

\newcommand{\duster}{DUSt3R\xspace}

\newcommand{\bp}{\boldsymbol{p}}
\newcommand{\bq}{\boldsymbol{q}}
\newcommand{\bu}{\boldsymbol{u}}

\newcommand{\argmin}{\operatornamewithlimits{argmin}}

\newcommand{\R}[1]{{%
    \textbf{%
        \ifstrequal{#1}{1}{\textcolor{red}{R#1}}{%
        \ifstrequal{#1}{2}{\textcolor{blue}{R#1}}{%
        \ifstrequal{#1}{3}{\textcolor{magenta}{R#1}}{%
        \ifstrequal{#1}{4}{\textcolor{teal}{R#1}}{%
                           \textcolor{cyan}{R#1}%
        }}}}%
    }%
}}

\makeatletter
\newcommand*{\addFileDependency}[1]{
  \typeout{(#1)}
  \@addtofilelist{#1}
  \IfFileExists{#1}{}{\typeout{No file #1.}}
}
\makeatother

\makeatletter
\renewcommand{\paragraph}{%
  \@startsection{paragraph}{4}%
  {\z@}{-0.5em}{-0.5em}%
  {\normalfont\normalsize\bfseries}%
}
\makeatother

\newcommand{\method}{DualPM\xspace}
\newcommand{\methods}{DualPMs\xspace}

\newcommand\rurl[1]{%
  \href{https://#1}{\nolinkurl{#1}}%
}

\def\paperID{5469}
\def\confName{CVPR}
\def\confYear{2025}
\def\paperTitle{\method: Dual Posed-Canonical Point Maps \\
for 3D Shape and Pose Reconstruction}
\def\authorBlock{
    Ben Kaye\textsuperscript{1}\thanks{Equal contribution} \qquad
    Tomas Jakab\textsuperscript{1}$^*$ \qquad
    Shangzhe Wu\textsuperscript{2,3} \qquad
    Christian Ruprecht\textsuperscript{1} \qquad
    Andrea Vedaldi\textsuperscript{1} \\[0.4em]
    \textsuperscript{1}University of Oxford \quad
    \textsuperscript{2}Stanford University \quad
    \textsuperscript{3}University of Cambridge\\[0.1em]
    {\small\rurl{dualpm.github.io}}
}

\title{\paperTitle}
\author{\authorBlock}

\begin{document}
\twocolumn[{
\maketitle
\begin{center}

\includegraphics[width=\linewidth,clip,trim=0.1cm 0.4cm 1.0cm 0.2cm]{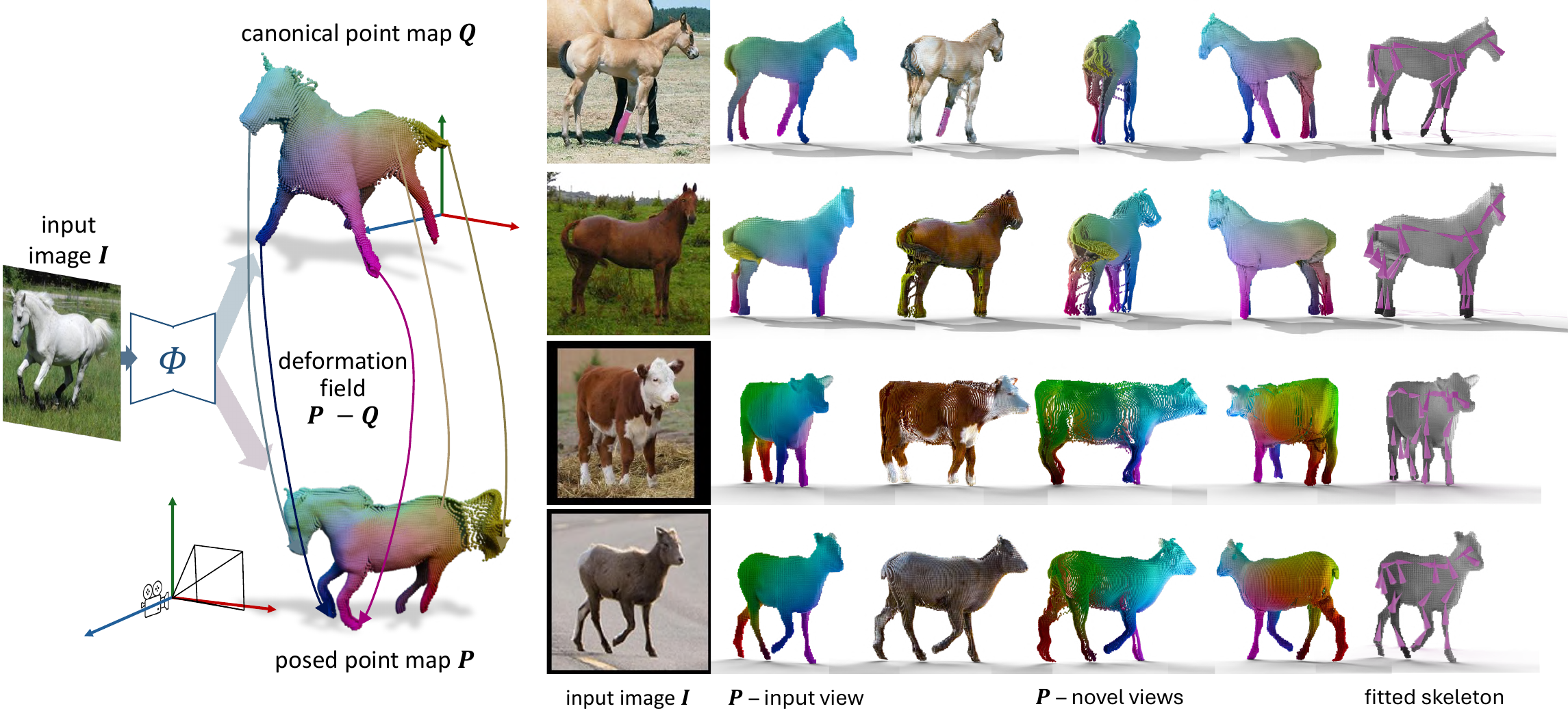}
\end{center}
\vspace{-1em}
\captionsetup{type=figure}
\captionof{figure}{
\emph{Left}:  
We map an image of an object to its \textbf{Dual Point Maps} ({\methods}), a pair of point maps $P$, defined in a camera space, and $Q$, defined in a canonical space where the object has a neutral pose.
The pose is thus given by the flow $P-Q$.
\emph{Right}:
The \methods are easy to predict with a neural network, enabling effective 3D object reconstruction and facilitating geometric tasks like detecting 3D keypoints and fitting a 3D skeleton.
For visualization, we color each point with its coordinate in the canonical point maps.
}%
\label{fig:teaser}
\vspace{1em}
}]

\def\thefootnote{*}\footnotetext{Equal contribution.}\def\thefootnote{\arabic{footnote}}

\begin{abstract}
The choice of data representation is a key factor in the success of deep learning in geometric tasks.
For instance, \duster recently introduced the concept of viewpoint-invariant point maps, generalizing depth prediction and showing that all key problems in the 3D reconstruction of static scenes can be reduced to predicting such point maps.
In this paper, we develop an analogous concept for a very different problem: the reconstruction of the 3D shape and pose of deformable objects.
To this end, we introduce Dual Point Maps (\method), where a pair of point maps is extracted from the \emph{same} image—one associating pixels to their 3D locations on the object and the other to a canonical version of the object in its rest pose.
We also extend point maps to amodal reconstruction to recover the complete shape of the object, even through self-occlusions.
We show that 3D reconstruction and 3D pose estimation can be reduced to the prediction of \methods.
Empirically, we demonstrate that this representation is a suitable target for deep networks to predict.
Specifically, we focus on modeling quadrupeds, showing that \methods can be trained purely on synthetic 3D data, consisting of one or two models per category, while generalizing effectively to real images.
With this approach, we achieve significant improvements over previous methods for the 3D analysis and reconstruction of such objects.
\end{abstract}

\section{Introduction}%
\label{sec:intro}

An important question in 3D computer vision is finding the optimal way of interfacing visual geometry and neural networks.
One approach is to use neural networks for pre-processing, for instance, to detect and match image keypoints.
Once these 2D primitives are extracted, visual geometry can be used to infer the 3D structure of the scene by solving a system of equations or via optimization~\cite{wang24vggsfm:}.
The alternative approach is to task the neural network with outputting \emph{directly} the geometric quantities of interest, such as depth maps~\cite{yang24depthv2} or the camera pose~\cite{wang23posediffusion:}.

Until recently, the general consensus was that inferring accurate 3D information should be, whenever possible, left to visual geometry, leaving to neural networks to fill gaps such as feature matching and monocular prediction.
However, this view is increasingly challenged as researchers find better ways of reducing geometric tasks to the calculation of quantities that can be predicted very effectively by neural networks.
Most recently, \duster~\cite{wang24dust3r:,duisterhof24mast3r-sfm:} has demonstrated the power of predicting \emph{point maps}, namely images that associate each pixel with its corresponding 3D point in the scene.
They show that many tasks in the reconstruction of static scenes, such as matching, camera estimation, and triangulation, can be solved trivially from the point maps predicted by a neural network.

In this work, we ask whether a similar intuition applies to a different class of problems, namely the monocular reconstruction of the \emph{3D shape and pose of deformable objects}.
We do so by developing a new network-friendly representation, which we call \emph{Dual Point Maps} (\methods).

Our setting is a substantial departure from \duster, and so are their point maps and our \methods.
Nevertheless, the starting point is the same:
a point map in both \duster and \method associates each image pixel with its corresponding 3D point on the object.
Predicting this point map gives us a 3D reconstruction of the object, which is useful but provides no information about the \emph{pose} of the object.

To clarify this further, consider the example of a horse.
Its pose is \emph{defined} as the deformation that takes the horse from its neutral pose to the pose observed in the image.
Hence, reconstructing its pose means finding the transformation, or deformation field, between two versions of the 3D object:
the one seen in the image and the \emph{canonical} version of the same object in a neutral pose.
Knowledge of a single point map gives no information about this transformation.
So the question is: How can we extend point maps so that recovering this deformation field becomes trivial?

To solve this problem, our \methods predicts not one, but \emph{two} point maps from the \emph{same image} (\cref{fig:teaser}).
They map each pixel to \emph{two} versions of the same 3D point.
The first version is the usual reconstruction of the object in 3D.
The second version, however, is the \emph{location of the point in the canonical version of the object}.
With this representation, it is easy to recover the pose of the object: in fact, the deformation field that we wish to recover is simply the \emph{difference} between the two point maps.
Knowledge of this deformation field is, in turn, useful for various analysis tasks, such as recovering 3D keypoints or estimating the object's articulation in terms of a skeleton (see~\cref{fig:teaser}).

\begin{figure}[t]\centering
\includegraphics[width=\columnwidth]{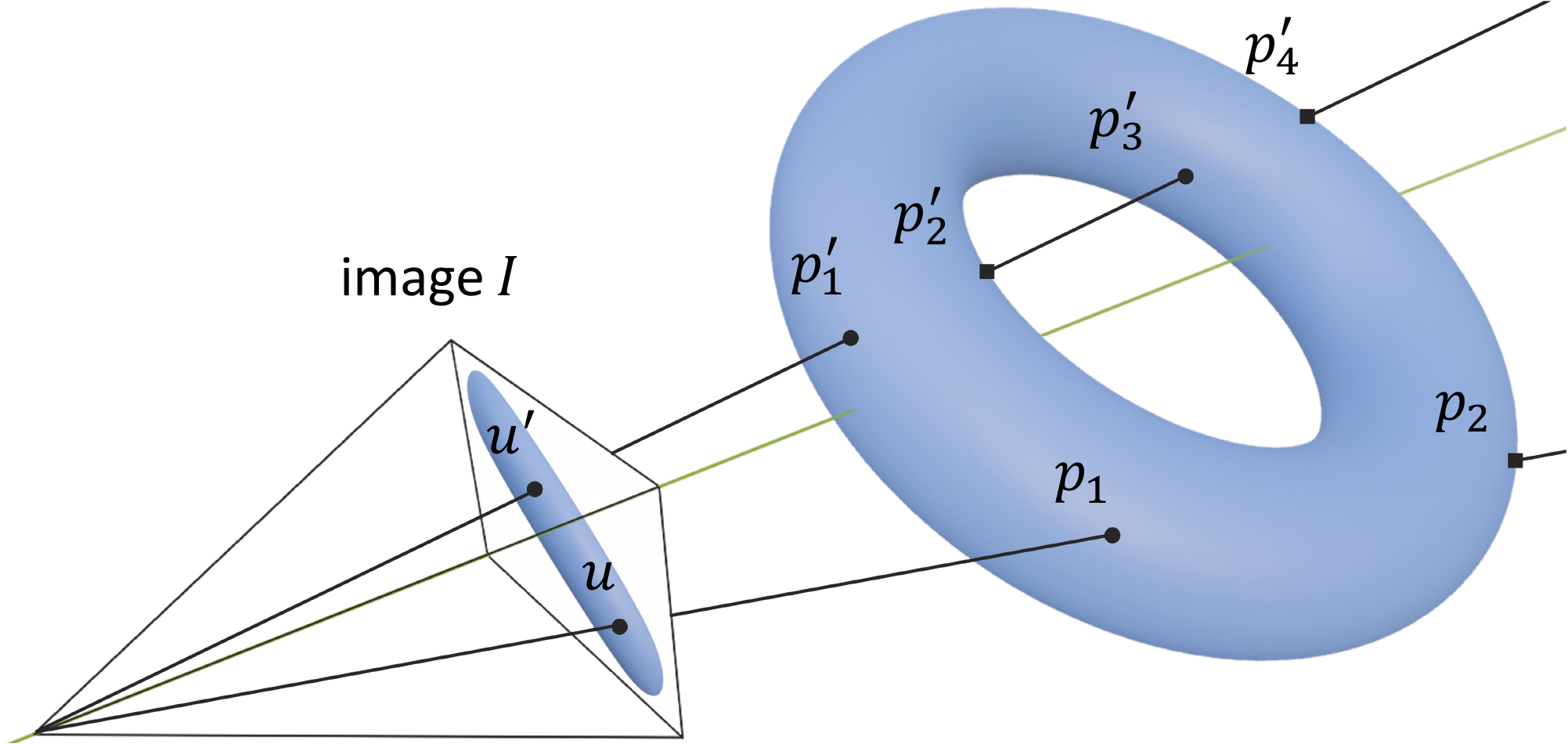}
\caption{An \emph{amodal point map} associates each pixel $\bu$ to an even number of objects points $\bp_i$, corresponding to the locations where the pixel's ray intersects the object's surface.
Predicting an amodal point map reconstructs the entire object despite self-occlusions.}\label{fig:amodal_pm}
\end{figure}

The canonical version of the point map is also \emph{viewpoint invariant} and bears some resemblance to the Normalized Object Coordinates of~\cite{wang19normalized} and the Canonical Surface Maps of~\cite{thewlis17dense,kulkarni19canonical}.
As such, its prediction is similar to a keypoint labeling problem, which can be solved effectively by a neural network.
Furthermore, we show that the canonical point map is also a good feature map in its own right; in fact, our model predicts the posed point map from the canonical one, using it as a feature extractor instead of, for instance, DINOv2~\cite{oquab24dinov2:}.

A shortcoming of point maps, in general, is that they only reconstruct the \emph{visible} part of the object.
To address this issue, we introduce the idea of estimating \emph{amodal} point maps; namely, each pixel is mapped not only to the visible point of the object but to \emph{all} points on the object that project onto that pixel, regardless of self-occlusions (\cref{fig:amodal_pm}).
To effectively represent such point maps for neural network prediction, we propose a layered representation.
In this representation, the first layer encodes the visible points, and each subsequent layer captures the next set of points occluded by the previous layer.
This concept is similar to depth peeling, a standard technique used in rasterization in computer graphics.
With this extension, we can provide a full reconstruction of the object's shape and pose.

To test our ideas, we consider the problem of reconstructing the 3D shape and pose of quadrupeds from monocular images.
Quadrupeds represent a widely studied category of deformable objects~\cite{kulkarni20articulation-aware,wu21dove:,wu23magicpony,jakab24farm3d:,li24learning}, with synthetic 3D data readily available for this class~\cite{jakab24farm3d:}, enabling controlled evaluations.
Furthermore, we show that synthetic data derived from one or two 3D models per category is sufficient to train a \method model that generalizes robustly to real-world data thanks to our effective representation, which makes learning relatively easy.
With this, we demonstrate that we can improve previous methods by a large margin for the 3D analysis and reconstruction of this type of object, even on out-of-domain real-world datasets, both in terms of correspondence and 3D reconstruction.
We also show the ability of the point maps to reconstruct the pose of the object in terms of their deformation fields and suggest applications such as fitting articulated 3D skeletons to the object, which can be used for motion transfer and animations.

To summarize, our contributions are as follows:  
\begin{enumerate}
    \item We introduce the novel concept of \method, reducing monocular 3D shape and pose reconstruction to predicting a \emph{pair} of point maps: one in canonical space and the other in posed camera space.  
    \item We extend point maps to enable amodal reconstruction with a layered representation, producing the complete 3D shape of articulated objects.  
    \item We demonstrate that \methods can be effectively predicted by neural networks, with synthetic data sufficing for training.  
    \item We show that tasks like 3D reconstruction, keypoint transfer, deformation estimation, and skeleton fitting reduce to predicting \methods.
    \item We achieve significant improvements over prior methods for 3D analysis and reconstruction of quadrupeds, even when trained on minimal synthetic data.
\end{enumerate}
\section{Related work}%
\label{sec:related}

\paragraph{Deformable reconstruction.}

Our work is related to the reconstruction of deforming scenes.
Many such works consider the \emph{optimization} approach, where a 4D model of the scene is fitted to a video sequence.
In works like
Video Pop-up~\cite{russell14video},
Monocular 3D~\cite{kumar17monocular}, and 
MonoRec~\cite{wimbauer21monorec:},
the 3D reconstruction of different video frames is regularized, for instance, to enforce approximate local rigidity.

More often, the 4D scene is represented as the deformation of a canonical configuration.
DynamicFusion~\cite{newcombe15dynamicfusion:} is an early example of this approach.
More recent works like Neural Volumes~\cite{lombardi19neural} reconstruct both shape and appearance.
D-NeRF~\cite{pumarola21d-nerf:},
Neural Radiance Flow~\cite{du21neural},
Neural Scene Flow Fields~\cite{li21neural},
Dynamic Video Synthesis~\cite{gao21dynamic},
DynIBaR~\cite{li23dynibar:}, and
MorpheuS~\cite{wang23morpheus:}
extend NeRF~\cite{mildenhall20nerf:} with dense deformation fields to capture motion.
Similarly,
4D Gaussian Splatting~\cite{wu234d-gaussian},
Gaussian Flow~\cite{lin24gaussian-flow:},
DynMF~\cite{kratimenos23dynmf:},
GauFRe~\cite{liang23gaufre:},
MoSca~\cite{lei24mosca:},
Street Gaussians~\cite{yan24street},
Dynamic Gaussian Marbles~\cite{stearns24dynamic},
Shape of Motion~\cite{wang24shape},
Ego Gaussians~\cite{zhang24egogaussian:}, and
4DGS~\cite{wu234d-gaussian}
also estimate deformations, but in the context of 3D Gaussian Splatting~\cite{kerbl233d-gaussian} rather than NeRF.

Rather than modeling dense deformation fields, approaches like
Dynamic NeRF~\cite{gafni21dynamic},
Seeing 3D Objects~\cite{sharma22seeing},
K-planes~\cite{fridovich-keil23k-planes:}, and 
HexPlane~\cite{cao23hexplane:}
add time or pose parameterization to the NeRF model.
Approaches like
4D Visualization~\cite{bansal204d-visualization} and
NVSD~\cite{yoon20novel} predict novel views of a dynamic scene without explicit 3D reconstruction.

Fourier PlenOctrees~\cite{wang22fourier},
NeRFPlayer~\cite{song23nerfplayer:}, and
RT-4DGS~\cite{yang24real-time}
focus on the problem of storing and rendering 4D models efficiently.
4DGen~\cite{yin1234dgen:},
Align Your Gaussians~\cite{ling24align},
DreamScene4D~\cite{chu24dreamscene4d:}, and
L4GM~\cite{ren24l4gm:}
consider the problem of generating 4D models from text (text-to-4D).
PhysGaussian~\cite{xie23physgaussian:},
MD-Splatting~\cite{duisterhof23md-splatting:},
Gaussian Splashing~\cite{feng24gaussian}, and
VR-GS~\cite{jiang24vr-gs:}
consider reconstruction and generation of 4D scenes that account for physical principles.

Works like Neural Human Video Rendering~\cite{liu20neural},
AutoAvatar~\cite{bai22autoavatar:},
Neural Body~\cite{peng21neural},
Dynamic Facial RF~\cite{shen22learning},
Relighting4D~\cite{chen22relighting4d:},
Animate124~\cite{zhao23animate124:},
Dynamic Gaussian Mesh~\cite{liu24dynamic}, and
IM4D~\cite{lin23im4d:}
specialize in the reconstruction of articulated characters like humans and animals.
Often, these define the object in a canonical neutral pose and encode the pose as a deformation thereof.

\paragraph{Learning deformable 3D object categories.}

Works like
3D Menagerie~\cite{zuffi173d-menagerie:},
Unsupervised 3D~\cite{wu20unsupervised},
LASR~\cite{yang21lasr:},
BANMo~\cite{yang22banmo},
Dessie~\cite{li24dessie:},
DOVE~\cite{wu21dove:},
LASSIE~\cite{yao22lassie:},
MagicPony~\cite{wu22magicpony:},
Farm3D~\cite{jakab24farm3d:},
3D-Fauna~\cite{li24learning}, and 
others~\cite{Goel2020ucmr,zuffi2018lions,yang2023rac,yao23artic3d:,ruegg2022barc} 
have considered the problem of learning models of 3D object categories.
In these models, the pose is expressed as the object deformation with respect to a version of the object in a neutral pose, which usually takes the form of a category-level template.
Such a template can be learned either in a weakly-supervised manner~\cite{wu21dove:,yao22lassie:,wu22magicpony:,jakab24farm3d:,li24learning,yang2023rac,yao23artic3d:} or from 3D data, most notably SMPL~\cite{loper15smpl:}, SMAL~\cite{biggs18creatures}, and more recently, VAREN~\cite{zuffi2024varen}.
Concurrent Dessie also introduces a synthetic data generation pipeline similar to ours.

\paragraph{Point maps and canonical maps.}

DUSt3R~\cite{wang24dust3r:} was the first paper to intuit the power of predicting point maps instead of depth for 3D reconstruction.
The follow-up MASt3R~\cite{duisterhof24mast3r-sfm:} builds on this to obtain a state-of-the-art Structure-from-Motion (SfM) system.
MonST3R~\cite{zhang24monst3r:} extends DUSt3R to monocular video, reconstructing dynamic point maps in a fixed reference frame.

The concept of learning a map that sends pixels to canonical object points can be found in the dense equivariant mapping of~\cite{thewlis17dense,thewlis17unsupervised,thewlis18modelling}.
Normalized Object Coordinates (NOCs)~\cite{wang19normalized} modify this idea for rigid objects by sending pixels to 3D points defined on the object's surface in a canonical pose.
Our approach also does this but with two key differences:
(1) the input object is deformable rather than rigid, and
(2) we predict dual point maps, which encode the object deformation beyond a simple rigid transformation.

\begin{figure*}[tp]
    \centering
    \includegraphics[width=\linewidth, trim=0 0.1cm 0 0.1cm, clip]{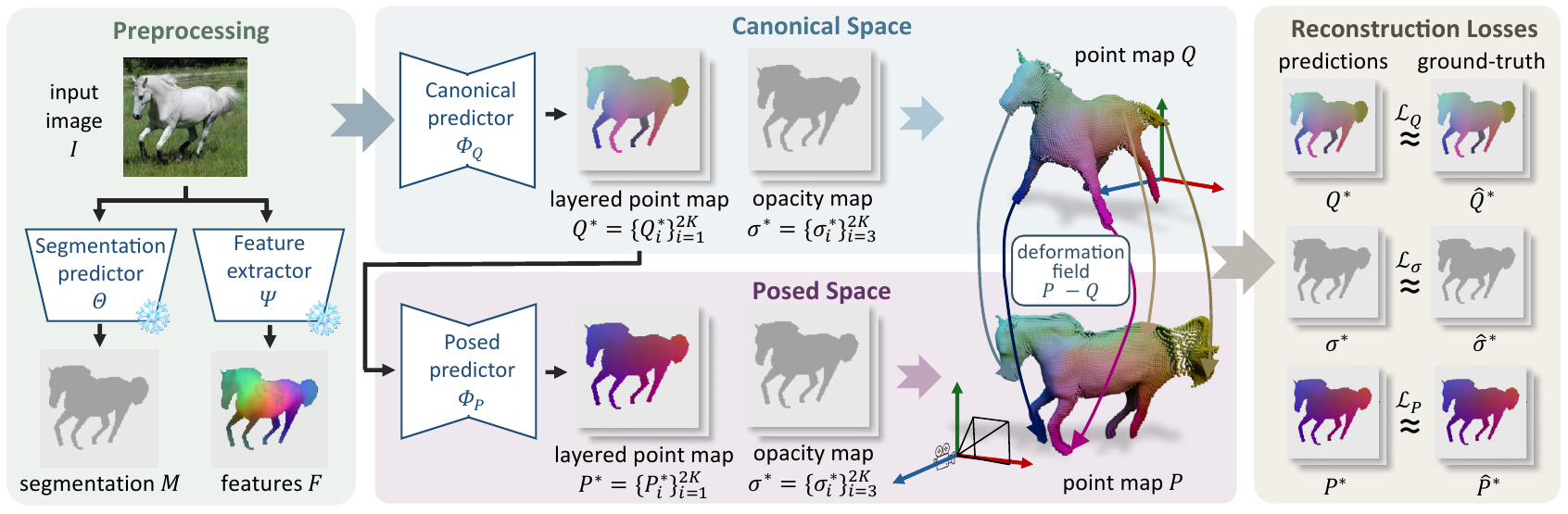}
    \caption{\textbf{Method overview.}
    We preprocess the input image $I$ by obtaining a segmentation mask $M$ and image features $F$ using pretrained networks.  
    Subsequently, we predict the layered canonical point map $\{Q_i^*\}_{i=1}^{2K}$.  
    Conditioned on $\{Q_i^*\}_{i=1}^{2K}$, we predict the layered posed point map $\{P_i^*\}_{i=1}^{2K}$, along with the associated opacity maps $\{\sigma_i^*\}_{i=3}^{2K}$.  
    Both layered point maps are then decoded to produce the canonical point map $Q$ and the posed point map $P$.  
    The training process is supervised using the ground truth point maps and opacity maps.
    }%
    \label{fig:pipeline}
\end{figure*}

\section{Method}%
\label{sec:method}

Let $I\in\mathbb{R}^{3\times H\times W}$ be an image containing a deformable object we wish to reconstruct in 3D, along with its pose.
We assume that a mask $M\in \{ 0,1 \}^{H\times W}$ is available, indicating the object's location in the image.
Next, we introduce the concept of \emph{dual point maps} (\cref{sec:dual-pointmaps}) and its \emph{amodal} extension (\cref{sec:amodal-pointmaps}).

\subsection{Dual point maps}%
\label{sec:dual-pointmaps}

A \emph{point map} $P \in \mathbb{R}^{3\times H\times W}$ is a tensor containing a 3D point for each pixel of the input image $I$.
We are particularly interested in the pixels $\bu \in M$ contained within the object's mask $M$.
Here, $\bu$ is the image of some point $\bp$ that belongs to the surface of the object, known as the \emph{imaged point}.
The image sample $I(\bu)$ is the color of the imaged point (an RGB triplet), and $\bp = P(\bu) \in \mathbb{R}^3$ is the 3D location of the point, expressed in the reference frame of the camera.

By estimating the point map $P$ from the image $I$, we reconstruct the visible portion of the 3D object.
This is similar to predicting a depth map but provides strictly \emph{more} information.
In fact, the 3D point \emph{does not} depend on the camera's intrinsic parameters, such as the focal length, as it is directly expressed in the camera's reference frame.
As shown in~\cite{wang24dust3r:}, this fact can be used to \emph{infer} the camera intrinsics.
However, this point map does not capture the object's deformation and thus cannot be used by itself to recover its articulated pose.

We address this limitation by introducing the novel concept of \emph{Dual Point Maps} (\methods).
In addition to $P$, we predict a second point map $Q \in \mathbb{R}^{3\times H\times W}$ defined in \emph{canonical space}.
By ``canonical'', we mean two things:
first, the points in $Q$ are expressed in the object reference frame rather than in the camera frame; second, the object is in a canonical pose, usually a neutral pose.
Hence, unlike $P$, the point map $Q$ is invariant to the object's pose and deformation.

The pair of point maps $P$ and $Q$ implicitly encode the object deformation: they tell us that pixel $\bu$ corresponds to a point that is located at $\bq = Q(\bu)$ in the rest pose and at $\bp = P(\bu)$ once posed.
This provides rich information about the object's pose.
For example, given two images $I_1$ and $I_2$ of the object in two different poses, we can match a pixel $\bu_1$ in the first image to its corresponding pixel $\bu_2$ in the second by matching the corresponding canonical points $Q_1(\bu_1)$ and $Q_2(\bu_2)$, \ie,
$$
\bu_2(\bu_1) = \argmin_{\bu_2 \in M_2} \| Q_1(\bu_1) - Q_2(\bu_2) \|^2.
$$
This works because the canonical points are view- and pose-invariant.
Then, we can infer the so-called `scene flow', or motion between the 3D points, as
$
P_2(\bu_2(\bu_1)) - P_1(\bu_1).
$

\paragraph{Predicting the dual point maps.}

To predict the dual point maps from the image $I$, we introduce a predictor neural network
$
(P,Q) = \Phi(I).
$
Its architecture is illustrated in \cref{fig:pipeline}.
A key property of the canonical point map $Q$ is its \emph{invariance} to the object's pose and deformation.
This invariance simplifies its prediction by a neural network, as it reduces the task to a pixel labeling problem.
To further ease the prediction of $Q$ and improve generalization, we leverage recent advances in self-supervised image feature learning~\cite{zhang23a-tale,oquab24dinov2:,esser21taming}, and condition its prediction on strong features $F$ extracted from the image $I$ using a pre-trained network $\Psi$, $F = \Psi(I)$ from~\cite{zhang23a-tale}.
These features serve as a good proxy for the canonical point map, as they have been shown to be nearly invariant to the object's pose and deformation, thereby simplifying the network's task.
We then condition the prediction of $P$ on $Q$ instead of $F$, as this improves the model's generalization on out-of-distribution images, as demonstrated in \cref{sec:ablation}.
This shows that learning canonical point maps as an intermediate representation benefits the 3D reconstruction task.
To summarize, we first predict $Q$ using the features $F$, and then condition the prediction of $P$ on $Q$:
$$
Q = \Phi_Q(\Psi(I)), \quad P = \Phi_P(Q).
$$

We extend both networks $\Phi_Q$ and $\Phi_P$ to also predict a \emph{per-pixel confidence score}.
For example, the map $\Phi_P$ outputs the map $c_P(\bu) > 0$, which is used to train the model using the self-calibrated L2 loss from~\cite{novotny18capturing,kendall17what,wang24dust3r:}:
$$
\mathcal{L}_P
=
\frac{1}{|M|}
\sum_{\bu \in M} c_P(\bu)
\| \hat{P}(\bu) - P(\bu) \|^2
-
\alpha \log c_P(\bu),
$$
where $\alpha > 0$ is a constant and $\hat{P}$ is the ground-truth point map.
An analogous loss $\mathcal{L}_Q$ is used for $\Phi_Q$.

\subsection{Amodal point maps}%
\label{sec:amodal-pointmaps}

A limitation of dual point maps is that they only capture the visible portion of the object.
We now show how to predict a complete 3D point cloud for the object, including points that are not visible due to \emph{self-occlusion}.

To this end, we introduce the concept of an \emph{amodal point map}.
Namely, we associate each pixel $\bu \in M$ within the object mask to the sequence of 3D points $(\bp_1,\bp_2,\dots)$ that intersect the camera ray through $\bu$ in order of increasing distance from the camera center.
Hence, $\bp = \bp_1$ is the imaged point considered in \cref{sec:dual-pointmaps}, $\bp_2$ is the point that would be imaged if a hole were plucked in the object surface at $\bp_1$, and so on.
The goal is to predict the \emph{amodal point map} $\mathcal{P}$, defined as the map that associates each pixel $\bu$ with the sequence of 3D points $\mathcal{P}(\bu) = (\bp_1,\bp_2,\dots)$.

The length of the sequence $\mathcal{P}(\bu)$ is not fixed but depends on the pixel $\bu$.
The sequence length is always even because the ray must exit the object once it enters it, as we assume the camera is positioned outside the object.
To account for this, we extend the network $\Phi_P$ to first predict a pair of points $(\bp_1, \bp_2)_1$ for each pixel $\bu$.
Subsequently, additional sets of outputs $(\bp_{2k-1}, \bp_{2k})_k$ are added for $k \in [2, K]$, where $K$ denotes the total number of pairs, capturing further potential ray intersections with the object.
Unlike the first set of intersections, these additional intersections may not always occur; therefore, we also predict opacities $(\sigma_{2k-1}, \sigma_k)_k$ for these cases.
As a result, each pixel $\bu$ is mapped to $3 \times 2 + (3 + 1) \times 2 \times (K - 1)$ scalar outputs.
The network $\Phi_Q$ is extended similarly.
The losses $\mathcal{L}_P$ and $\mathcal{L}_Q$ are also extended to account for the amodal point maps.
Additionally, we add a loss $\mathcal{L}_\sigma$ to supervise the opacity predictions.

\paragraph{Layered point maps.}

To efficiently represent the amodal point map for neural network prediction, we introduce an image-based \emph{layered} representation for \methods, denoted as $(P^*, Q^*)$.
In this representation, the first layer encodes the visible points, and each consecutive layer captures the next set of points occluded by the previous layer.
We also represent the opacity maps $\sigma^*$ in the same layered fashion.

Since we supervise our method with synthetic data, we can easily generate training targets for this representation.
Each vertex of the object mesh is assigned a 6D attribute, consisting of the posed vertex's 3D position and the corresponding vertex's 3D position in canonical space.
The mesh is then rendered from the camera's viewpoint.
During the rendering process, the rasterizer generates a set of pixel-aligned layers, $L_i \in \mathbb{R}^{6 \times H \times W}$, which contain the rendered 6D attributes, together with the associated opacity map $\hat{\sigma}_i^* \in [0, 1]^{H \times W}$ for every ray-object intersection.
We iterate through the rasterized layers from the front to the back until the number of layers specified for our model is reached.
This approach yields ground-truth \methods with a number of layers corresponding to the ray-object intersections.
We split the attributes of each layer corresponding to the canonical and posed positions, yielding ground-truth layered canonical point maps
$$  
\hat{Q}^* = (\hat{Q}_1^*, \hat{Q}_2^*, \dots, \hat{Q}_N^*), \quad \hat{Q}_i^* \in \mathbb{R}^{3 \times H \times W},  
$$  
and layered posed point maps  
$$  
\hat{P}^* = (\hat{P}_1^*, \hat{P}_2^*, \dots, \hat{P}_N^*), \quad \hat{P}_i^* \in \mathbb{R}^{3 \times H \times W},  
$$  
with their associated opacity maps $\sigma_i^*$.
The number of layers $N$ corresponds to $2K$.
The canonical $Q$ and posed $P$ point maps can then be extracted from this representation by masking out transparent points.

\subsection{Training}%
\label{sec:training-data}
\begin{figure}[t]\centering
\includegraphics[width=\columnwidth]{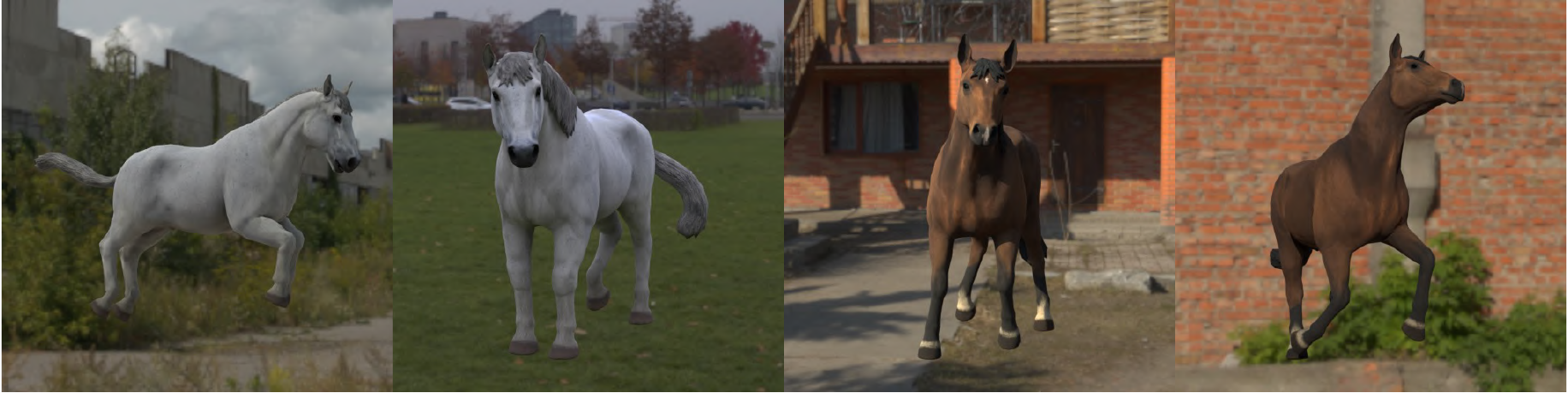}
\caption{\textbf{Synthetic training data.}
We generate synthetic training data by rendering a rigged 3D model of a category in various poses with different environmental maps and under random viewpoints.
}\label{fig:synth}
\end{figure}

We train a separate model for each category using synthetic data generated from one or two articulated 3D assets per category sourced from the Animodel dataset~\cite{jakab24farm3d:}.
We use a separate model for each biological sex in cases where animals exhibit distinct anatomical structures specific to males and females, such as horns in cattle.
Following the data generation pipeline described in Animodel, we randomly sample object poses, camera poses, and illumination conditions.
These are modeled by randomly sampling from a collection of environment maps.
For each sample, we render the object into an image $I$ and generate the corresponding image-based layered paired point maps $P^*$ and $Q^*$, which serve as targets for the model, using the procedure detailed in \cref{sec:amodal-pointmaps}.
We illustrate our synthetic training data in \cref{fig:synth}.
\section{Experiments}%
\label{sec:results}

    \begin{figure*}[tp]
\centering
\begin{minipage}{0.99\textwidth}
\begin{minipage}{0.19\textwidth}
    \centering
    \caption*{}
\end{minipage}%
\hspace{0.05cm}
\begin{minipage}{0.28\textwidth}
    \centering
    \caption*{3D-Fauna~\cite{li24learning}}
\end{minipage}%
\hspace{0.1cm}
\begin{minipage}{0.48\textwidth}
    \centering
    \caption*{Ours}
\end{minipage}
\vspace{-0.1cm}
\includegraphics[width=\textwidth, trim=6mm 2mm 3mm 4mm, clip]{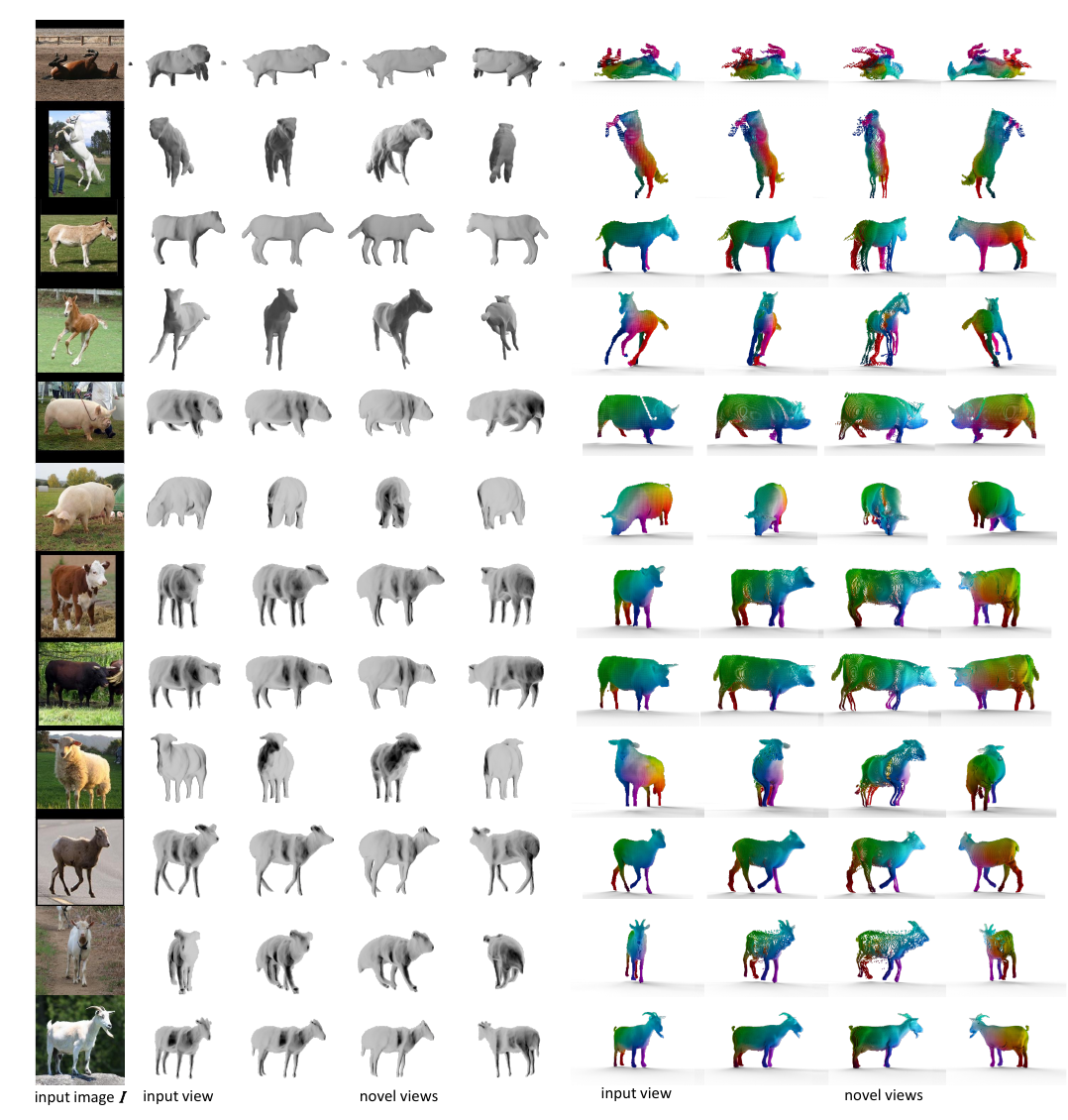} %
\end{minipage}
\caption{\textbf{Comparison with the state-of-the-art.}
We compare our method with 3D-Fauna~\cite{li24learning}.
For visualization, we paint each point of the recovered point map $P$ with its canonical coordinate.  
Our reconstructed shape aligns more closely with the object's structure compared to 3D-Fauna and adapts well to large variations within the object category.
Our method reliably recovers shapes outside its training distribution, such as foals and donkeys (rows 3 and 4).
It also performs well on very challenging poses (rows 1 and 2), where 3D-Fauna completely fails.}%
\label{fig:qual_comparison}
\end{figure*}
\begin{table*}[]
\setlength{\tabcolsep}{5.5pt}
\centering
\small
\newcommand{\xpm}[1]{\small {$\pm\ #1$}}
\resizebox{1\textwidth}{!}{
\begin{tabular}{lrrrrrrrrr}
    \toprule
    \multirow{2}{*}{Method} & \multicolumn{3}{c}{PCK (\%)} & \multicolumn{3}{c}{RMS Chamfer Distance (cm)} & \multicolumn{3}{c}{Model-view RMS Chamfer Distance (cm)} \\
    \cmidrule(r){2-4} \cmidrule(r){5-7} \cmidrule(r){8-10}
    & \multicolumn{1}{c}{Horse} & \multicolumn{1}{c}{Cow} & \multicolumn{1}{c}{Sheep} & \multicolumn{1}{c}{Horse} & \multicolumn{1}{c}{Cow} & \multicolumn{1}{c}{Sheep} & \multicolumn{1}{c}{Horse} & \multicolumn{1}{c}{Cow} & \multicolumn{1}{c}{Sheep} \\
    \midrule
    A-CSM \cite{kulkarni20articulation-aware} & 32.9 & 26.3 & 28.6 & 11.75 \xpm{3.83} & 9.52 \xpm{2.41} & 9.24 \xpm{2.40} & 38.13 \xpm{13.89} & 33.51 \xpm{11.52} & 29.04 \xpm{9.35} \\
    MagicPony \cite{wu23magicpony}            & 42.9 & 42.5 & 41.2 & 11.19 \xpm{3.08} & 10.29 \xpm{2.08} & --- & 20.82 \xpm{13.04} & 25.39 \xpm{13.43} & --- \\
    Farm3D \cite{jakab24farm3d:}              & 49.1 & 40.2 & 36.1 & 11.34 \xpm{3.22} & 9.63 \xpm{2.02} & 11.01 \xpm{1.87} & 29.52 \xpm{15.73} & 21.34 \xpm{12.85} & 21.52 \xpm{9.84} \\
    3D-Fauna \cite{li24learning}              & 53.9 & --- & --- & 11.86 \xpm{3.03} & 10.54 \xpm{2.26} & 9.61 \xpm{2.15} & 15.70 \xpm{6.82} & 14.08 \xpm{4.20} & 12.24 \xpm{3.17} \\
    Trellis \cite{xiang2024structured}       & --- & --- & --- & 6.93 \xpm{4.13} & 6.80 \xpm{3.24} & 5.91 \xpm{2.93} & 36.82 \xpm{16.02} & 26.54 \xpm{13.14} & 26.56 \xpm{12.06} \\
    \midrule
    DualPM (Ours)                             & \textbf{73.2} & \textbf{66.86} & \textbf{66.82} & \textbf{4.30} \xpm{1.50} & \textbf{3.18} \xpm{1.06} & \textbf{3.30} \xpm{1.20} & \textbf{5.49} \xpm{1.75} & \textbf{4.03} \xpm{2.11} & \textbf{4.22} \xpm{2.18} \\
    \bottomrule
\end{tabular}
}

\caption{\textbf{Quantitative evaluation.}  
We evaluate on PASCAL VOC, reporting PCK@0.1 (higher is better $\uparrow$), and on Animodel-Points, a derivative of Animodel ~\cite{jakab24farm3d:} for evaluating point clouds, reporting the Root-mean-square-error of bi-directional Chamfer Distance in centimeters (lower is better $\downarrow$). Point clouds are registered using the iterative closest points algorithm with bidirectional chamfer distance as its cost function. We report both the results with and without rotation alignment.
Our model, trained solely on data from one or two models per category, outperforms other state-of-the-art approaches, and has superior camera alignment.
}%
\label{tab:quant}
\end{table*}

We evaluate our \method in terms of both qualitative (\cref{sec:qualitative}) and quantitative performance (\cref{sec:quantitative}) across several quadruped categories.
We demonstrate that our \method can be used to fit 3D skeletons for motion transfer and animations (\cref{sec:skeleton}), assess the impact of our design choices in an ablation study (\cref{sec:ablation}), and showcase zero-shot generalization to unseen quadruped categories.
Additionally, we discuss the limitations of our method and provide technical and implementation details in the supplementary material.

\subsection{Qualitative evaluation}%
\label{sec:qualitative}

We qualitatively evaluate our method on challenging in-the-wild images of quadrupeds, specifically the horse, cow, sheep, and pig categories from the PASCAL VOC dataset~\cite{everingham2010pascal} and goat images from~\cite{billah2020goat}.
The results are presented in~\cref{fig:teaser,fig:qual_comparison}.
Our method successfully recovers the 3D shape and pose of quadrupeds from a single image, despite significant variations in pose, shape, and appearance within each animal category.
For instance, in the horse category, although the method is trained on a single adult horse model, it generalizes well to foals and dwarf horses.
We also compare our method with the state-of-the-art approach for single-view reconstruction of deformable objects, 3D-Fauna~\cite{li24learning}, and present side-by-side comparisons in~\cref{fig:qual_comparison}.
Our method performs better in recovering the shape and pose than 3D-Fauna, which struggles to accurately follow the object's structure and fails to reconstruct intricate details specific to each object.

\subsection{Quantitative evaluation}%
\label{sec:quantitative}

Following prior work on deformable object reconstruction~\cite{kulkarni20articulation-aware,li20self-supervised,wu23magicpony,li24learning}, we evaluate our method on the keypoint transfer task using the PASCAL VOC dataset~\cite{everingham2010pascal}.
We adhere to the protocol of~\cite{kulkarni20articulation-aware} and report the percentage of correctly transferred keypoints with a 10\% threshold (PCK@0.1) between pairs of images.
The results are presented in~\cref{tab:quant}, where we compare our method with state-of-the-art approaches.

We also measure the bi-directional Chamfer distance between the predicted point maps and the ground truth 3D shape on the Animodel dataset introduced in Farm3D~\cite{jakab24farm3d:}.
This dataset comprises renderings of 3D models of horses, cows, and sheep in various poses.
To ensure a fair comparison, we train a separate model on data that excludes the poses used in this benchmark.
The performance of our method and others is reported in~\cref{tab:quant}\footnote{Amendment: the Animodel evaluation was not suitable for evaluating point clouds, and we have updated our table with an improved benchmark. The original published table is available in the appendix in~\cref{tab:animodel_supp_old}. Benchmark details are available in \cref{asec:benchmark}.}

Our method consistently outperforms all other approaches, including self-supervised methods trained on large datasets of real images of quadruped categories (3D-Fauna~\cite{li24learning}) and approaches trained on large-scale 3D asset datasets (Trellis~\cite{xiang2024structured}), demonstrating its ability to generalize to unseen poses and shapes.
Moreover, it exhibits generalization to real-world images of the same category, despite being trained solely on synthetic data obtained from one or two templates per category.
Additionally, it demonstrates some zero-shot generalization across categories, as shown in the supplementary material.

\subsection{Skeleton fitting and animation}%
\label{sec:skeleton}
\begin{figure}[t]\centering
\includegraphics[width=\linewidth, trim=0.0cm 0.2cm 0.8cm 0cm, clip]{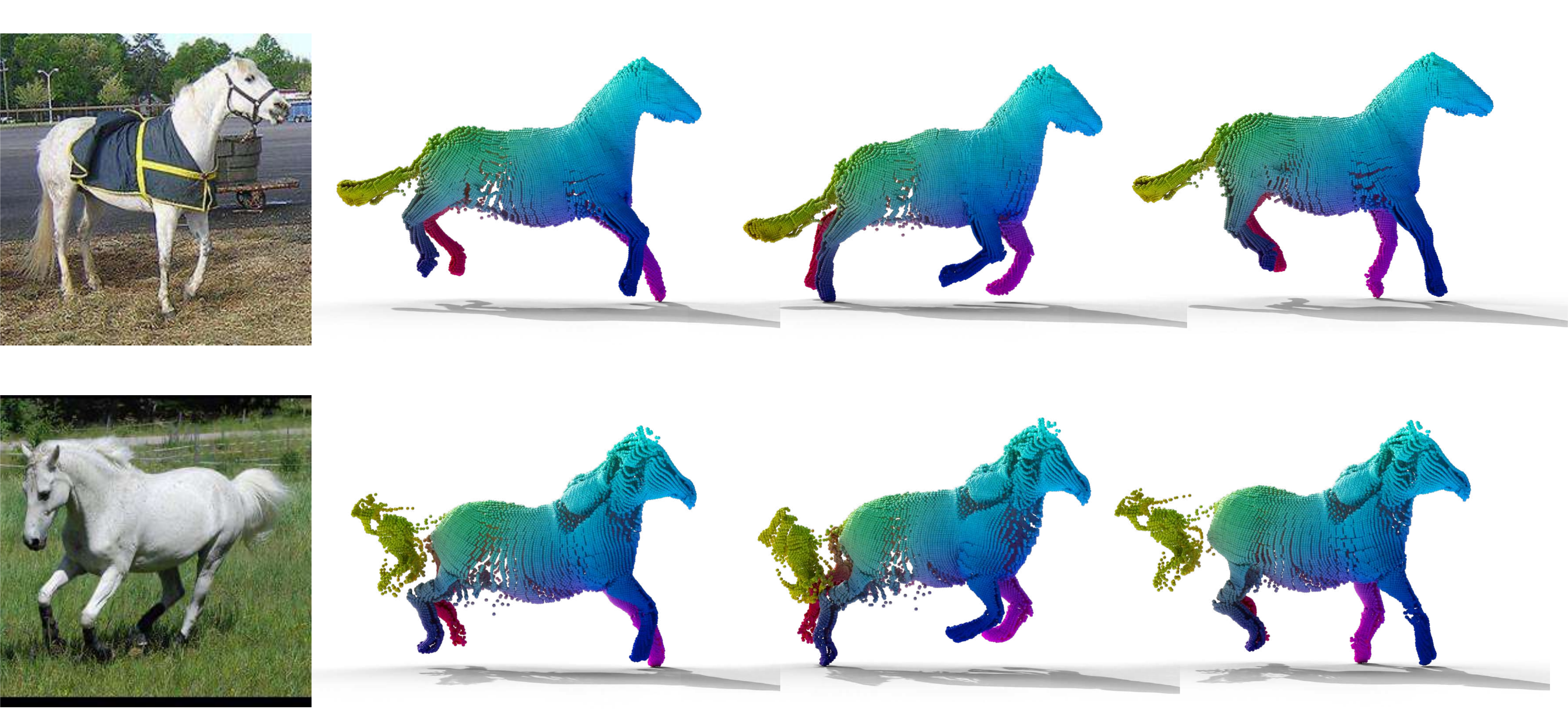}
\caption{
\textbf{Animation.}  
\methods allow for fitting a 3D skeleton, which can subsequently be used to animate the 3D reconstruction by utilizing existing animations through motion retargeting.
}\label{fig:anim}
\end{figure}

With a predicted \method $(P, Q)$, we can easily fit an instance-specific 3D skeleton.  
Our method is trained on synthetic data from a model rigged with a skeleton composed of bones $B$ that deform the mesh.
These bones are defined in the canonical space, and each bone is associated with a set of vertices on the mesh through skinning weights.

For each point $(\bp_i, \bq_i)$ of the \method $(P, Q)$, we identify the closest vertex of the mesh in the canonical space to the canonical point $\bq_i$, then associate the point $(\bp_i, \bq_i)$ with the skinning weights of that vertex.
For each bone $b \in B$, we identify subsets $\{(\bp_i, \bq_i)_b\}$, where the associated skinning weights between the bone $b$ and the points $(\bp_i, \bq_i)$ are higher than 0.5.
We then compute the transformation that aligns the subset of canonical points $\{(\bq_i)_b\}$ to the subset of posed points $\{(\bp_i)_b\}$ using the Procrustes algorithm.
This transformation is applied to the bone $b$ in the canonical space to derive its corresponding pose in the posed space.
This process is repeated for all bones in the skeleton to fit the 3D skeleton to the posed points.
We show examples of our fitted skeletons in~\cref{fig:teaser}.

The fitted skeleton can then be used to animate our 3D reconstructions through motion retargeting from existing animations defined on the rigged model used for training.
We demonstrate examples of this in~\cref{fig:anim}.

\subsection{Ablation study}%
\label{sec:ablation}

We conduct an ablation study to evaluate the impact of different components of our model on its performance and report the results in~\cref{tab:ablation}.
Specifically, we examine the advantages of predicting pixel-aligned point maps over directly predicting bone rotations, which are typically used to articulate a rigged mesh model in prior approaches~\cite{wu23magicpony,jakab24farm3d:,li24learning}.
To test this, we train a version of the method proposed in~\cite{wu23magicpony,li24learning} on our synthetic dataset, where the bone rotation regressor is supervised directly with ground-truth rotations.

The results show that this model performs worse than our proposed approach.
We hypothesize that this disparity arises because our representation is more amenable to neural networks.
Predicting pixel-aligned point maps effectively transforms the task into a pixel-labeling problem, which is simpler for neural networks to learn.
In contrast, directly predicting a chain of 3D rotations for a rigged model is inherently more complex.

\begin{figure}[t]\centering
\includegraphics[width=0.97\columnwidth,clip,trim=0cm 0.4cm 1.4cm 0cm]{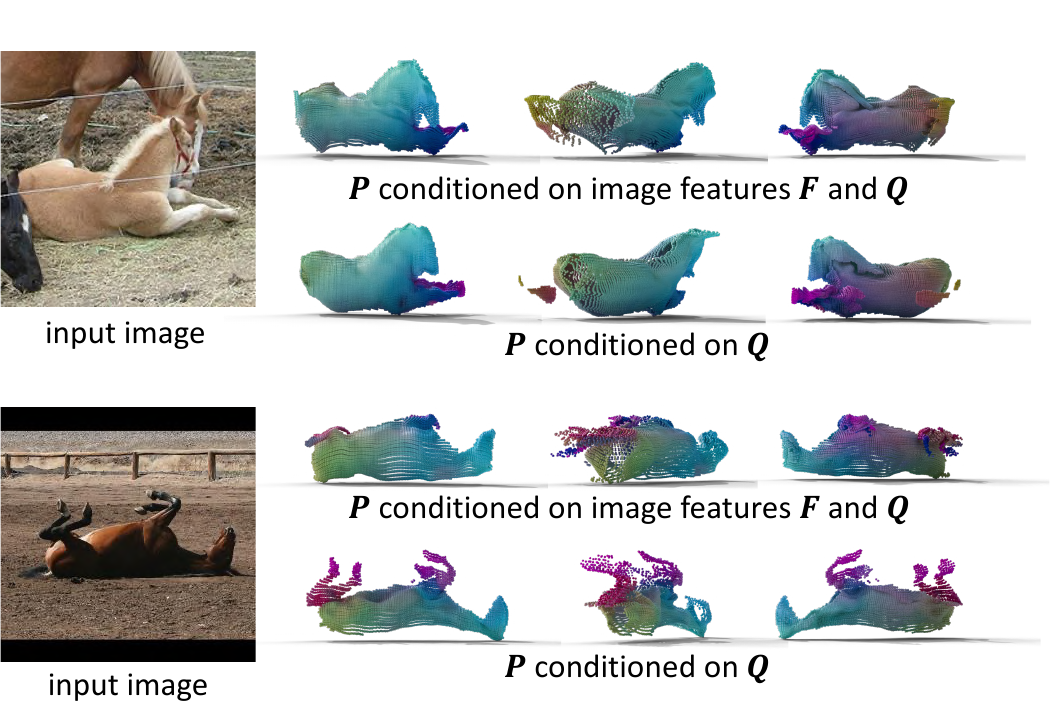}
\caption{\textbf{Effect of different conditioning schemes.}
Conditioning solely on the canonical point maps $Q$, as opposed to using both the image features $F$ and the canonical point maps $Q$, results in better generalization, particularly in extreme out-of-distribution images.
}\label{fig:ablations}
\end{figure}

Additionally, we investigate the effect of conditioning the predictions of the posed point maps $P$ on the input image features $F$ and the canonical point maps $Q$.
When the model is conditioned on the image features $F$, it tends to overfit to the more variable features of the training dataset.
In contrast, conditioning the posed point maps $P$ solely on the canonical point maps $Q$ leads to better generalization, particularly on out-of-distribution images, as we also demonstrate qualitatively in~\cref{fig:ablations}.

Finally, we analyze the impact of varying the number of layers used in the layered amodal point map predictions.
The performance on the Animodel benchmark shows minimal change, as self-occlusions beyond four occluded surfaces are less frequent in the dataset.
The number of predicted layers can be easily adjusted to match the complexity of the data.

\begin{table}[htbp]
    \centering
    \small

    \label{tab:ablation}
    \begin{tabular}{lr}
        \toprule
        Method & {RMS Chamfer Distance (cm) $\downarrow$} \\
        \midrule
        Predicting bone rotations \\ 
        (supervised 3D-Fauna \cite{li24learning}) & 6.79 $\pm$ 2.47 \\
        \midrule
        \multicolumn{2}{l}{\textit{Our method}} \\
        \quad 8-layers & 4.45 $\pm$ 1.61 \\
        \quad 4-layers & 4.30 $\pm$ 1.50 \\ %
        \midrule
        \multicolumn{2}{l}{\textit{Ablations on 4-layer model}} \\
        \quad $P$ conditioned on $F$ and $Q$ & 4.54 $\pm$ 1.60 \\
        \quad $P$ conditioned on $F$ & 4.48 $\pm$ 1.46 \\
        \bottomrule
    \end{tabular}
    \caption{
        \textbf{Ablation study on Animodel-points horses category.}
        We report the root-mean-square bi-directional Chamfer Distance in centimetres (lower is better $\downarrow$) for the horse category of Animodel-points.
        The evaluation follows the protocol as~\cref{tab:quant}.
    }
\end{table}

\section{Conclusion}%
\label{sec:conclusion}

We presented \methods, a novel representation for monocular 3D shape and pose reconstruction of deformable objects.  
Predicting \methods is straightforward for a neural network, and they enable solving a variety of geometric tasks for deformable objects, including not only 3D shape and pose reconstruction but also 3D keypoint localization, skeleton fitting, and motion transfer.
Despite being trained solely on synthetic data generated from one or two models per category, our method generalizes effectively to real images and outperforms state-of-the-art methods on the task of 3D shape and pose reconstruction of horses, cows, and sheep.

\paragraph*{Acknowledgments.}

We thank Paul Engstler for helpful discussions.
Ben Kaye is supported by the EPSRC Centre for Doctoral Training in Autonomous Intelligent Machines and Systems, EP/S024050/1.
Tomas Jakab and Andrea Vedaldi are supported by ERC-CoG UNION 101001212.
Tomas Jakab is also supported by VisualAI EP/T028572/1.

{
\small
\bibliographystyle{ieeenat_fullname}
\bibliography{vedaldi_general,vedaldi_specific,11_references}
}

\ifarxiv
\clearpage
\appendix

\renewcommand{\thefigure}{{A}\arabic{figure}}
\renewcommand{\thetable}{{A}\arabic{table}}
\renewcommand{\theequation}{{A}\arabic{equation}}

\section*{Appendix}

\section{Generalization to unseen categories}%

\begin{figure}[b]\centering
    \includegraphics[width=\linewidth, trim=0.0cm 0.2cm 0.8cm 0cm, clip]{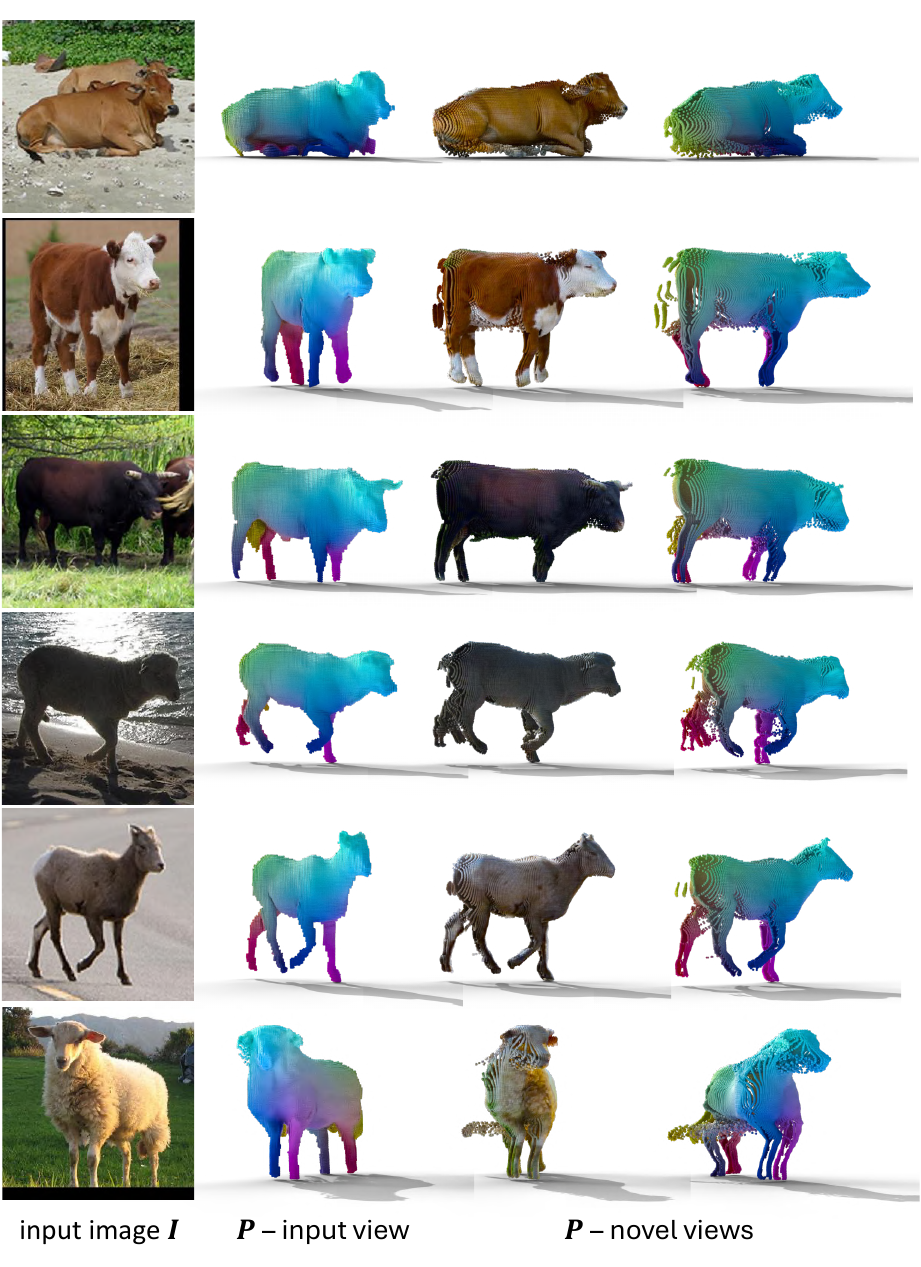}
    \caption{
    \textbf{Results on unseen categories.}  
    A version of our model trained only on the horse category also demonstrates robust generalization to the unseen categories such as cow and sheep, despite being trained solely with a single horse model.  
    }\label{fig:add_results}
\end{figure}

\label{sec:generalization}
Given the generalization capabilities of our method demonstrated within a single category, we analyze the generalization of a model trained on a single category to unseen categories.
Specifically, we consider a model trained on horses and evaluate its performance on cow and sheep categories.
We evaluate our approach using the same datasets as for the horses, PASCAL VOC~\cite{everingham2010pascal} and Animodel~\cite{jakab24farm3d:}, following the same evaluation protocol, with results reported in~\cref{tab:animodel_supp}.
Furthermore, we provide additional qualitative results on the same dataset in~\cref{fig:add_results}.
Our method exhibits strong zero-shot generalization to these categories, outperforming state-of-the-art approaches on both datasets, despite being trained exclusively on a single horse model.

\begin{table*}[]
\centering
\small
\newcommand{\xpm}[1]{\small {$\pm\ #1$}}
\begin{tabular}{lrrrrrr}
    \toprule
    \multirow{3}{*}{Method} & \multicolumn{2}{c}{PCK (\%)} & \multicolumn{4}{c}{Chamfer Distance (cm)}  \\
    \cmidrule(r){2-3} \cmidrule(r){4-7}
    & \multicolumn{1}{c}{\multirow{2}{*}{Cow}} & \multicolumn{1}{c}{\multirow{2}{*}{Sheep}} & \multicolumn{2}{c}{Real-Sized}  & \multicolumn{2}{c}{Normalized}  \\
    \cmidrule(r){4-5} \cmidrule(r){6-7}
    & & & \multicolumn{1}{c}{Cow} & \multicolumn{1}{c}{Sheep} & \multicolumn{1}{c}{Cow} & \multicolumn{1}{c}{Sheep} \\
    \midrule
    A-CSM~\cite{kulkarni20articulation-aware} & 26.3 & 28.6 & 6.71 \xpm{1.81} & 2.84 \xpm{0.77} & 2.35 \xpm{0.68} & 2.48 \xpm{0.70} \\
    MagicPony~\cite{wu23magicpony}            & 42.5 & 41.2 & 7.22 \xpm{1.53} & 3.43 \xpm{0.73} & 2.53 \xpm{0.59} & 3.00 \xpm{0.68} \\
    Farm3D~\cite{jakab24farm3d:}              & 40.2 & 36.1 & 6.91 \xpm{1.49} & 3.79 \xpm{0.55} & 2.41 \xpm{0.54} & 3.31 \xpm{0.49} \\
    3D-Fauna~\cite{li24learning}              & ---  & ---  &  9.19 \xpm{2.40}            & 3.51 \xpm{0.88}            &  3.20 \xpm{0.80}            &  3.06 \xpm{0.76}            \\
    \midrule
    Ours                                      & \textbf{63.0} & \textbf{64.2} & \textbf{4.74} \xpm{1.40} & \textbf{2.32} \xpm{0.78} & \textbf{1.67} \xpm{0.55} & \textbf{2.03} \xpm{0.71} \\
    \bottomrule
\end{tabular}
\caption{\textbf{Evaluation on unseen cow and sheep categories.}  
We evaluate on PASCAL VOS, reporting PCK@0.1 (higher is better $\uparrow$), and on Animodel~\cite{jakab24farm3d:}, reporting the bi-directional Chamfer Distance in centimeters (lower is better $\downarrow$).  
Our model, trained solely data from a single horse model, outperforms state-of-the-art approaches, which were trained on data that included these specific categories.  
}
\label{tab:animodel_supp}
\end{table*}

\section{Limitations}%
\label{sec:limitations}
\begin{figure}[b]\centering
\includegraphics[width=\columnwidth, trim=0.3cm 0 0.3cm 0, clip]{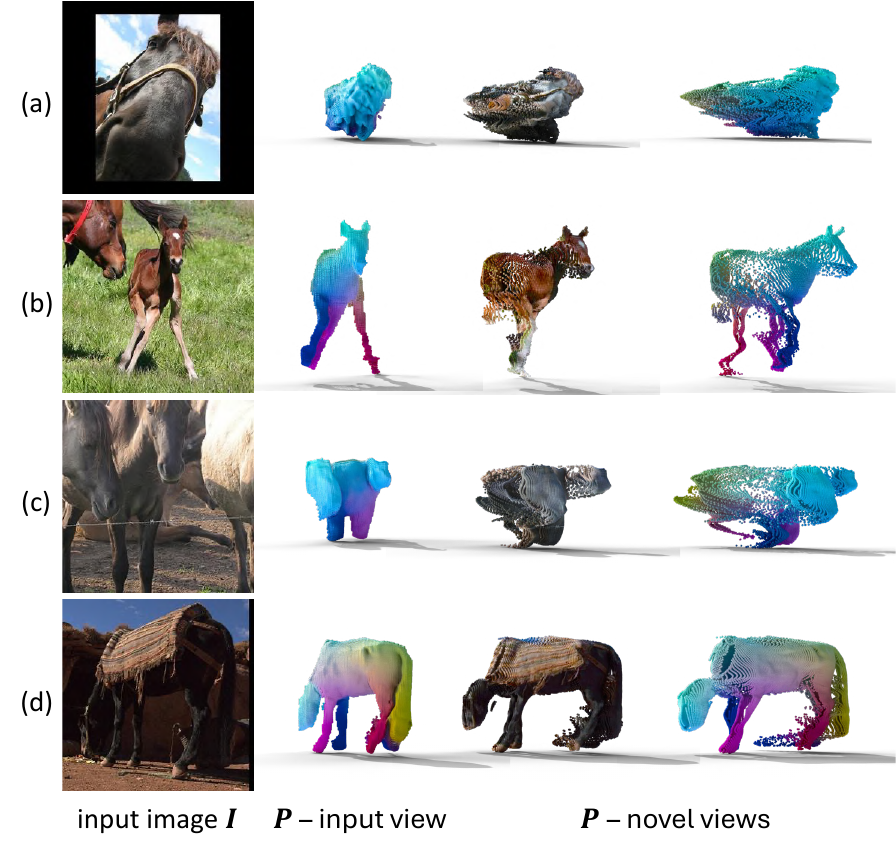}
\caption{
\textbf{Typical failure cases.} We illustrate representative failure cases caused by (a) extreme viewpoints, (b) shapes and poses far from the training distribution, and (c-d) inaccuracies in the object segmentation masks.
}\label{fig:fail}
\end{figure}

Despite demonstrating surprising generalization, a current limitation of our method is that any additional synthetic 3D models added to the training dataset would have to be in the same canonical space as the training data. 
Addressing the challenging problem of aligning the canonical spaces of multiple 3D models would allows us to train on significantly larger datasets which could in turn lead to significant gains in performance for our proposed model.
Another limitation of our method is that it is not specifically trained to handle occlusions caused by other objects.
This is a limitation shared with other methods, such as 3D-Fauna~\cite{wu23magicpony,li24learning}.
To address this, we plan to extend the data generation pipeline to include synthetic occlusions.
Additionally, as the 3D reconstruction problem is often ambiguous for the unseen parts of objects, our method predicts only the expectation over all possible reconstructions, which can lead to unrealistic results for the invisible regions.
We illustrate our typical failure cases in~\cref{fig:fail}.

\section{Technical details}%
\label{asec:details}

\paragraph{Network architecture.}
We obtain the segmentation mask $M$ using the Segment Anything method~\cite{kirillov23segment}.
The feature extractor $\Psi$ is based on~\cite{zhang23a-tale} which combines pre-trained DINOv2~\cite{oquab24dinov2:} and StableDiffusion~\cite{rombach22high-resolution} networks.
Training image features are reduced to a 64-dimensional space using PCA following~\cite{wu22magicpony:}.
The dual point map predictors $\Phi_Q$ and $\Phi_P$ leverage a convolutional U-Net architecture based on~\cite{song21score-based}, comprising two blocks each and trained from scratch.
We predict $N = 4$ layers for layered amodal point maps as more have little effect on the performance (\cref{sec:ablation}), likely due to the low frequency of multiple self-occlusions in our datasets.
The number of layers can be easily increased should the data require it.
The output resolution of the layered point maps is set to 160$\times$160.

\paragraph{Training.}
We use the Adam optimizer~\cite{kingma15adam:} for training.
Our model is trained for 100k steps with a batch size of 12.
The learning rate is set to $6 \times 10^{-4}$, with a step scheduler applied, featuring a 30k-step period and a decay factor of 0.5.

\paragraph{Training dataset.} 
The training dataset consists of approximately 30k rendered images per category.
We generate these images using a single rigged model per animal species.
For cow, sheep, and goat, we use a separate model for each sex category, incorporating major sex-specific attributes such as horns.
Each model includes up to three different textures and 50 animated actions, such as running, walking, and drinking.
We also randomly sample from a pool of 742 HDRI environmental maps to provide diverse lighting conditions for the training images.
We then randomly sample camera viewpoints and poses from the animated actions to generate the training images.
\Cref{fig:synth} showcases the horse model and some of the generated images used for training.

\section{Benchmark details}
\label{asec:benchmark}

We introduce \textbf{Animodel-Points}, a benchmark specifically designed to address the limitations of using the original Animodel benchmark for evaluating point cloud generation models. Previous evaluations on Animodel were suboptimal for two primary reasons: the benchmark was designed for \textbf{meshes}, not point clouds, and its evaluation backend (\texttt{trimesh.registration})\footnote{Trimesh: A Python library for loading and using triangular meshes. \url{https://github.com/mikedh/trimesh}} only optimizes a one-sided Chamfer distance.

Animodel-Points resolves these issues by establishing a new protocol that directly evaluates point clouds using a more robust metric. The key improvements are:
\begin{itemize}
    \item \textbf{Model-view coordinates:} Targets are defined in model-view coordinates up to a depth offset, which allows for the additional evaluation of camera alignment performance.
    \item \textbf{Bidirectional Chamfer distance:} The evaluation protocol minimizes the mean squared bidirectional Chamfer distance, aligning the optimization objective with the reported cost for a fairer assessment.
    \item \textbf{Standardized point cloud processing:} A clear pre-processing and evaluation pipeline ensures consistency and comparability of results.
\end{itemize}

\subsection*{Methodology}

The methodology is divided into two stages: preparing the ground truth data and the protocol for evaluating a prediction.

\paragraph{Data pre-processing}
To convert the original Animodel meshes into the Animodel-Points format, we apply the following steps:
\begin{enumerate}
    \item \textbf{Coordinate transformation:} The source mesh is transformed into model-view coordinates.
    \item \textbf{Uniform scaling:} The mesh is uniformly scaled by a factor of $V^{-1/3}$, where $V$ is the mesh volume, to normalize its size.
    \item \textbf{Point sampling:} 20,000 points are uniformly sampled from the surface of the scaled mesh to serve as the ground truth target.
\end{enumerate}

\paragraph{Evaluation protocol}
For a given generated point cloud, the following evaluation steps are performed:
\begin{enumerate}
    \item \textbf{Resampling:} If the input point cloud does not contain 20,000 points, it is resampled to this size.
    \item \textbf{Rotational ambiguity handling:} The 20,000 point sample is duplicated, and the duplicate is rotated 180 degrees around its vertical axis.
    \item \textbf{Subsampling for fitting:} A subset of 10,000 points is sampled from the input to be used in the alignment process.
    \item \textbf{Alignment:} Both the original and rotated samples are aligned to the ground truth target using the Iterative Closest Point (ICP) algorithm. The process runs for a maximum of 200 steps or until convergence, using the MSE bidirectional Chamfer distance as the objective function.
    \item \textbf{Final score:} The lower of the two costs from the alignment steps is reported as the final score.
\end{enumerate}

Chamfer distance performance is evaluated by providing scale, rotation, and translation degrees of freedom to the transformation estimation. Model-view chamfer distance is evaluated by restricting the degrees of freedom to just scale and translation.

\paragraph{Published results}
\begin{table*}[]
\centering
\newcommand{\xpm}[1]{\small {$\pm\ #1$}}
\begin{tabular}{lrrrrrr}
    \toprule
    \multirow{3}{*}{Method} & \multicolumn{6}{c}{Chamfer Distance (cm)}  \\
    \cmidrule(r){2-7}
    & \multicolumn{3}{c}{Real-Sized}  & \multicolumn{3}{c}{Normalized}  \\
    \cmidrule(r){2-4} \cmidrule(r){5-7}
    & \multicolumn{1}{c}{Horse} & \multicolumn{1}{c}{Cow} & \multicolumn{1}{c}{Sheep} & \multicolumn{1}{c}{Horse} & \multicolumn{1}{c}{Cow} & \multicolumn{1}{c}{Sheep} \\
    \midrule
    A-CSM \cite{kulkarni20articulation-aware} & 7.60 \xpm{3.07} & 6.71 \xpm{1.81} & 2.84 \xpm{0.77} & 2.73 \xpm{1.13} & 2.35 \xpm{0.68} & 2.48 \xpm{0.70} \\
    MagicPony \cite{wu23magicpony}           & 7.19 \xpm{2.35} & 7.22 \xpm{1.53} & 3.43 \xpm{0.73} & 2.58 \xpm{0.80} & 2.53 \xpm{0.59} & 3.00 \xpm{0.68} \\
    Farm3D \cite{jakab24farm3d:}             & 7.65 \xpm{2.21} & 6.91 \xpm{1.49} & 3.79 \xpm{0.55} & 2.76 \xpm{0.83} & 2.41 \xpm{0.54} & 3.31 \xpm{0.49} \\
    3D-Fauna \cite{li24learning}             & 8.69 \xpm{2.38} & 9.19 \xpm{2.40} & 3.51 \xpm{0.88} & 3.13 \xpm{0.85} & 3.20 \xpm{0.80} & 3.06 \xpm{0.76} \\
    Trellis \cite{xiang2024structured}       & 5.11 \xpm{3.53} & 5.50 \xpm{3.37} & 2.23 \xpm{1.39} & 1.85 \xpm{1.31} & 1.93 \xpm{1.23} & 1.96 \xpm{1.22} \\
    \midrule
    Ours                                     & \textbf{3.13} \xpm{2.13} & \textbf{2.63} \xpm{1.23} & \textbf{1.52} \xpm{0.88} & \textbf{1.11} \xpm{0.73} & \textbf{0.92} \xpm{0.43} & \textbf{1.33} \xpm{0.78} \\
    \bottomrule
\end{tabular}
\caption{\textbf{Evaluation on Animodel.}  
We evaluate on Animodel~\cite{jakab24farm3d:}, reporting the bi-directional Chamfer Distance in centimeters (lower is better $\downarrow$). Ours include results from models trained on data from each category.
}
\label{tab:animodel_supp_old}
\end{table*}

Previously published results were reported on the Animodel benchmark and are available in this table \cref{tab:animodel_supp_old}.

\fi
\end{document}